# The Preposition Project

Ken Litkowski
CL Research
9208 Gue Road
Damascus, MD 20872
ken@clres.com

Orin Hargraves
5130 Band Hall Hill Road
Westminster, MD 21158
orinkh@carr.org

**Abstract**

Prepositions are an important vehicle for indicating semantic roles. Their meanings are difficult to analyze and they are often discarded in processing text. The Preposition Project is designed to provide a comprehensive database of preposition senses suitable for use in natural language processing applications. In the project, prepositions in the FrameNet corpus are disambiguated using a sense inventory from a current dictionary, guided by a comprehensive treatment of preposition meaning. The methodology provides a framework for identifying and characterizing semantic roles, a gold standard corpus of instances for further analysis, and an account of semantic role alternation patterns. By adhering to this methodology, it is hoped that a comprehensive and improved characterization of preposition behavior (semantic role identification, and syntactic and semantic properties of the preposition complement and attachment point) will be developed. The databases generated in the project are publicly available for further use by researchers and application developers.

## 1 Introduction

Characterization of preposition meanings is important for understanding the semantic relations between elements of a sentence. The difficulty of this task arises from their polysemy. Defining English prepositions in a native speaker dictionary is a thankless task: the senses are many and complexly interrelated; the frequency of prepositions requires the study of numerous examples; and their treatment in dictionaries may cause confusion and information overload, since there is little agreement in minor, and sometimes even in major sense divisions. The lexicographer may suspect that the effort is of little value: native speakers do not often consult dictionaries to learn what a preposition means.

When the dictionary entries are used as the basis for processing text, the tables are turned: the definition of prepositions is of vital importance, and can be an important resource. But the distinctions for processing by the human mind in dictionaries are not in all cases easily or efficiently processed by computer, and the dictionary alone may be insufficient.

The Preposition Project (TPP) is designed to provide a comprehensive characterization of preposition senses suitable for use in natural language processing. It is attempting to fine-tune the distinctions within and among prepositions in a native speaker dictionary (the *Oxford Dictionary of English*, 2003) by comparing and contrasting them with the treatment of prepositions in two other sources: the instances of prepositions that are functionally tagged in FrameNet, and the treatment of prepositions in a traditional English grammar (Quirk et al., 1985). This paper will survey the project and describe initial findings of prepositional behavior that have come to light through this exercise.

## 2 The Preposition Project

Each of 847 preposition senses for 373 prepositions (including phrasal prepositions) will be characterized with a semantic role name and the syntactic and semantic properties of its complement and attachment point. Each sense will be further described by (1) a link to its definition in the *Oxford Dictionary of English*, (2) its basic syntactic function and meaning as described in Quirk et al. (1985), (3) other prepositions filling a similar semantic role, (4) FrameNet frames and frame elements, (5) other syntactic forms in which the semantic role may be realized, and (6) its position in a network of prepositions. This basic information is provided in a spreadsheet for each preposition. All data generated during this project is freely available for use by other researchers and application developers.[1]

The primary source of data is the set of corpus instances from FrameNet: sentences tagged with semantic roles (frame elements) for each preposition. Since FrameNet was not constructed with prepositions in mind, the examination of frame elements using a preposition provides a corpus that considerably facilitates the construction of a high-quality preposition database. The assumption is that frames and frame elements will help elucidate the meanings of prepositions.

### 2.1 The Preposition Sense Inventory

The *Oxford Dictionary of English* (ODE, 2003) (and its predecessor, the *New Oxford Dictionary of English* (NODE, 1997)) was chosen as the source of the preposition sense inventory because of the clarity and organization of its senses and its reliance on corpus evidence. Litkowski (2002) identifies prepositions in NODE, including phrasal prepositions. As indicated there, 373 prepositions (listed in the appendix) and 847 preposition senses were identified. This set, with modifications as noted in Litkowski (2002), forms the basis for TPP's sense inventory.[2]

TPP's database does not include the definitions and examples from ODE, since that is the intellectual property of Oxford University Press. The database provides a key to these definitions, so that a specific sense can be identified and a user of this database can gain a further understanding of the meaning conveyed by each sense by referring to the dictionary itself.

### 2.2 Methodology for Sense Disambiguation of Preposition Instances in FrameNet

The initial focus of TPP is on the most common and most polysemous prepositions. These include the following 20 prepositions that have six or more senses: about (6), above (9), after (11), against (10), around (6), at (12), by (22), for (14), from (14), in (11), into (9), of (18), on (23), over (16), through (13), to (17), towards (6), under (16), with (16), and within (6).

After selecting a preposition for study,[3] the FrameNet corpus instances are obtained using CL Research's publicly available FrameNet Explorer (FNE).[4] The FrameNet database includes approximately 7,500 XML lexical unit files, each containing tagged sentences for a specific lexical item and frame (e.g., the item

---

[1] http://www.clres.com/prepositions.html

[2] As noted above, any sense inventory may be fraught with problems and several have already come to light. It is hoped that the suitability of this sense inventory for NLP applications, particularly issues of granularity, can be investigated more thoroughly with the type of data being generated.

[3] **By**, **through**, **with**, and **for** have been completed as of March 19, 2005.

[4] http://www.clres.com/FNExplorer.html

**Table 1. Preposition Instance File Sample Lines**

| Frame | Frame Element | Lexical Unit | Subcorpus | Identifier-Position |
|---|---|---|---|---|
| Achieving_first | No_instances | originate.v | V-570-s20-np-ppby | |
| Arrest | Authorities | arrest.v | V-730-s20-ppby | 875350-43 |
| Arrest | Authorities | arrest.v | V-730-s20-ppby | 875353-71 |
| Arrest | Authorities | arrest.v | V-730-s20-ppby | 875362-160 |
| Arrest | No_instances | apprehend.v | V-730-s20-ppby | |

*move.v* in the **Motion** frame). Tagged sentences are grouped into subcorpora, each of which has a name. The name encodes syntactic properties of the subcorpus, e.g., **V-730-s20-ppacross**; which includes sentences using the verb *move* that include a prepositional phrase beginning with *across* (tagged as a **Path** frame element within the **Motion** frame).

FNE generates a text file of tagged FrameNet instances of a given preposition. FNE searches each lexical unit file to find subcorpora having **pp***prep* in the name. For each subcorpus having the target name (e.g., **ppby**), a line is written to the text file, containing: the frame name, the frame element, the lexical unit, the subcorpus name, and the sentence ID and starting position of the preposition in the sentence. Table 1 shows sample lines from the instance file for **by**.

In constructing a line, the FrameNet data for the sentence are examined to identify the frame element introduced by the target preposition. The example data indicate that no sentences containing a prepositional phrase beginning with **by** were tagged in the subcorpora for *originate.v* and *apprehend.v*, but that 3 sentences were tagged for *arrest.v* in the **Arrest** frame with the **Authorities** frame element. The file is initially sorted by frame name; imported into an Excel spreadsheet, it can be sorted on any other element. For **by**, 1314 lines were generated.[5]

Using this instance file as a guide, the lexicographer begins the process of analyzing the preposition's senses. A separate Excel spreadsheet is devised for the preposition, with one row for each sense (with the ODE sense number in parentheses). The lexicographer examines the definitions for the preposition, available information about the preposition in Quirk et al., and the FrameNet corpus instances. On the basis of this information, the lexicographer assigns an arbitrary and subjective semantic role name, intended to be a characterization of the sort of information that the given preposition introduces. He then identifies the usual syntactic function of a phrase with the preposition in the specific sense (*noun postmodifier (1); adverbial adjunct (2a), subjunct (2b), disjunct (2c), or conjunct (2d);* and/or *verb (3a) or adjective (3b) complement*, as described in paragraph 9.1, p. 657 of Quirk et al.). The lexicographer then ascertains the paragraph, if any, in Quirk et al. that provides a semantic description of the instant sense. This paragraph may also identify other prepostions that have a similar sense and use; these other prepositions are also recorded in the spreadsheet, along with any others that the lexicographer intuits may have a similar meaning.

Based on the definition and the corpus instances, the lexicographer then sets out to characterize the syntactic and semantic properties of the sense's complement and attachment point, based on an interpretation of the definition. These characterizations are preliminary and not based on any systematic criteria; however, this is not important at this stage of development. As described below, it is expected that these characterizations will be refined when disambiguation routines are developed. Table 2 shows this information for five (of 13) senses of **through**.[6]

---
[5]The instance file generated by this method does not represent all instances of a preposition in the FrameNet database.

Table 2. Sample Senses for 'through'

| Sense | Relation Name | Quirk Syntax | Quirk Paragraphs | Complement Properties | Attachment Properties |
|---|---|---|---|---|---|
| 1 (1) | ThingTransited | 2a, 3a | 9.25, 9.28 | opening, channel, or location | verbs of motion |
| 2 (1a) | ThingBored | 1, 2a, 3a | 9.25, 9.28 | permeable or breakable physical object | verbs denoting penetration |
| 3 (1b) | ThingTransited | 1, 2a, 3a | 9.25, 9.26, 9.28 | sth regarded as homogenous | verbs of motion |
| 4 (1c) | ThingPenetrated | 1, 2a, 3a | None | a permeable obstacle | a perceived object; sometimes complement of a verb of perception |
| 5 (1d) | ChannelTransited | 1, 2a, 3a | 9.19, 9.22, 9.27 | an opening or obstacle | copula or verb of location |

As indicated, the lexicographer assigns a sense number to each sentence instance. FNE is used for this purpose, by displaying all annotated instances of a lexical unit (such as *arrest.v*) entered on its search screen. In addition, all subcorpus names are displayed in a drop-down list; by selecting the relevant subcorpus (e.g., **V-730-s20-ppby**), the lexicographer can view just those sentences and determine which ODE sense of the preposition is applicable. Since similar items may be grouped together (i.e., frame name, frame element name, and lexical unit), several instances can be tagged at a time. Tagging about 1500 instances for a preposition takes about 10 hours.

The lexicographer may tag some instances with multiple senses. The lexicographer may also find, through the iterative exercise of examining FrameNet instances, that the sense division found in ODE does not quite match the reality of preposition use. In this case, additional lines may be created in the sense spreadsheet to accommodate new subsenses, or less frequently, entirely new senses. The lexicographer also keeps notes and prepares a summary describing the treatment of the preposition, noting any special or idiomatic uses of the preposition that may fall outside the defined sense inventory. Finally, the lexicographic description is compared with the Lexical Conceptual Structure inventory available from Dorr (1996) (9 senses for **by**, 2 senses for **through**, 15 for **with**, and 8 for **for**)

From a lexicographic perspective, it turns out that each source of information about the behavior of a preposition is incomplete in itself. All sources used in the project are complementary in providing an overall assessment of the meaning and characterization of the preposition. ODE may be found wanting when placed next to the FrameNet instances; this project may thus reveal further aspects of the appropriate sense inventory. ODE does not provide a summary picture of a preposition's meanings; the characterization in Quirk et al. provides such a perspective, but it too is incomplete, both in coverage of a particular meaning and in not identifying correspondences with other prepositions. The FrameNet database does not provide instances for all the senses. The Dorr inventory confirms the characterization here, but does not contain the same level of detail. Despite the (minor) deficiencies of each source, their combination appears to be quite comprehensive.

---

[6]The definitions in ODE for Table 2 are "(1) moivng in one side and out of the other side of (an opening, channel, or location): (a) so as to make a hole or opening in (a physical object); (b) moving around or from one side to the other within (a crowd or group); (c) so as to be perceived from the other side of (an intervening obstacle); (d) expressing the position or location of something beyond or at the far end of (an opening or an obstacle)."

Table 3. Frame:FrameElement Pairs Identified for Senses of 'through'

| Sense | Relation Name | Frame:FrameElement Pairs |
|---|---|---|
| 1 (1) | ThingTransited | Arriving:Path; Cause_motion:Path; Cotheme:Path; Departing:Path; Escaping:Location; Escaping:Path; Evading:Path; Fluidic_motion:Path; Mass_motion:Path; Motion:Path; Motion_directional:Path; Motion_noise:Path; Operate_vehicle:Path; Path_shape:Path; Placing:Goal; Placing:Path; Removing:Path; Roadways:Area; Self_motion:Area; Self_motion:Path; Breathing:Path |
| 2 (1a) | ThingBored | Cause_harm:Body_part; Impact:Impactee; Natural_features:Relative_location; Use_firearm:Path |
| 3 (1b) | ThingTransited | Emotion_heat:Location; Path_shape:Area; Ride_Vehicle:Path; Roadways:Path; Self_motion:Self_mover; Travel:Path |

## 3 Analyzing the Semantic Role for a Sense

With the tagged instances, a simple sort by sense number of the Excel spreadsheet identifies the (**Frame Frame_Element**) pairs for each sense. These pairs are aggregated into one list in the Sense Analysis spreadsheet (as shown in Table 3). As indicated above, the lexicographer identifies a semantic role label for each sense based on intuition. These labels are developed independently of (computational) linguistic theories and are mainly based on a general characterization of the sense information for the preposition. These labels are intended to be used in characterizing prepositional phrases, based on the criteria in the complement and attachment syntactic and semantic properties for disambiguating the prepositions. Gildea & Jurafsky (2002) developed a mapping of frame elements into 18 higher level semantic roles. The methodology followed here provides an alternative mapping that is more data-driven and less subjective.

In many senses for which FrameNet instances were identified, there is a clear correspondence between the frame element names and the semantic relation assigned by the lexicographer. But, they also show the range and variation of frame elements that have been developed by the FrameNet lexicographers. (**Frame FrameElement**) pairs and lexical units are shown in Table 4 for **through** (sense 3), given the label *ThingTransited*. This table suggests that this sense encapsulates a **Path** semantic role. Since other senses of **through** also have a **Path** role, the FrameNet lexicographer's assignment indicates a finer granularity on the type of path. The assignment of an **Area** frame element for *crisscross* suggests a finer granularity on the type of path, suggesting that the path might be *through* a region. This type of analysis demonstrates the richness of the data generated by tagging instances.

## 4 Refining Characterizations Through Disambiguation

In addition to the instances file, FNE also generates an XML file of the sentences themselves. These sentences (for which the preposition senses have been assigned) are suitable for the development of disambiguation routines for semantic role assignment. In this respect, these sentences are essentially equivalent to the lexical sample task followed in Senseval. In addition, since these instances are FrameNet tagged sentences, they provide a suitable dataset for the Senseval FrameNet semantic role task. (The XML files are available as part of TPP).

Table 4. Analysis of Sense 3 (ThingTransited) for 'through'

| Frame:Frame_Element | Lexical Units |
|---|---|
| *Emotion_heat:Location* | boil.v seethe.v burn.v |
| *Path_shape:Area* | crisscross.v |
| *Ride_Vehicle:Path* | hitchhike.v |
| *Roadways:Path* | bypass.n highway.n line.n motorway.n path.n pathway.n road.n street.n track.n trail.n |
| *Self_motion:Self_mover* | sprint.v |
| *Travel:Path* | journey.n journey.v tour.n travel.v |

Litkowski (2002) described a set of disambiguation tests for the preposition **of**, based solely on introspection of its definitions. Those tests are not sufficient. As implied in Table 2, the complement and attachment properties require a richer set of semantic tests for which suitable lexical resources do not presently exist. Sense 1 of **through** requires that the prepositional phrase be attached to a verb of motion; WordNet has a general *motion* category for verbs, so in this case, a suitable test can be made. However, for sense 2, it is necessary to identify verbs of penetration; no such category is available in WordNet. A Roget-style thesaurus might provide the necessary information (e.g., look up *penetration* in the thesaurus and then examine the verbs in the same thesaurus category).

The corpus instances developed in TPP will be used to refine the characterizations developed by the lexicographer. Disambiguation routines will be developed, particularly investigating the use of various lexical resources, such as WordNet, machine-readable dictionaries, and thesauruses. Many of the attachment characterizations suggest close ties to sets of verbs; development of appropriate disambiguation routines may reveal close associations with verb classes. This phase of TPP is not yet well developed.

## 5 Identifying Other Prepositions and Other Syntactic Realizations Filling the Same Semantic Roles

A tagged sentence in the FrameNet database identifies a specific frame element within a specific frame for the prepositional phrase introduced by the preposition. The frame element and frame can be used as a seed to find other ways recorded in FrameNet for realizing the combination. For example, as shown in Table 5,

Table 5. Variations in Syntactic Realizations of a Frame Element for 'by'

| Frame | Frame Element | Lexical Unit | GF | PT | Preposition |
|---|---|---|---|---|---|
| Arriving | Mode_of_transportation | arrive.v | Comp | PP | by |
| Arriving | Mode_of_transportation | arrive.v | Comp | PP | in |
| Arriving | Mode_of_transportation | come.v | Comp | PP | by |
| Arriving | Mode_of_transportation | return.n | Comp | PP | by |
| Arriving | Path | approach.v | Comp | PP | on |
| Arriving | Path | approach.v | Comp | PP | through |
| Arriving | Path | approach.v | Comp | PP | via |
| Arriving | Path | arrive.v | Comp | PP | through |
| Arriving | Path | arrive.v | Comp | PP | via |
| Arriving | Path | come.v | Comp | PP | round |
| Arriving | Path | come.v | Comp | PP | through |
| Arriving | Path | come.v | Comp | PP | via |
| Arriving | Path | come.v | Obj | NP | |
| Arriving | Path | enter.v | Comp | PP | at |
| Arriving | Path | enter.v | Comp | PP | by |
| Arriving | Path | enter.v | Comp | PP | through |
| Arriving | Path | enter.v | Comp | PP | via |
| Arriving | Path | get.v | Comp | PP | past |
| Arriving | Path | reach.v | Comp | PP | by |
| Arriving | Path | reach.v | Comp | PP | through |
| Arriving | Path | reach.v | Comp | PPing | |
| Arriving | Path | return.n | Comp | PP | towards |
| Arriving | Path | return.v | Comp | PP | across |

**by** introduces the frame element **Mode_of_transportation** or **Path** in the **Arriving** frame. FNE can be used to query the FrameNet database to determine other prepositions and other syntactic realizations in which these frame elements occur. The distinct patterns in which these occur are summarized by identifying all unique occurrences of (**Frame Frame_Element Lexical_Unit Grammatical_Function Phrase_Type Preposition**) within the database. (**Preposition** is included only when the **Phrase_Type** is **PP**.) There may be many sentences that have been tagged similarly, but only unique occurrences need to be identified to examine the distribution of the same frame element.

In Table 5, several combinations are evoked by the seed element. The **Mode_of_transportation** frame element was seeded by the instances for *arrive.v* and/or *come.v* (sense 8 of **by**); the **Path** element was evoked by the instances for *enter.v* (sense 5 of **by**). It can be seen that in addition to *by*, *in* is also used to indicate the **Mode_of_transportation** frame element, also as a **Complement** to the main verb. For the **Path** frame element, in addition to *by*, the prepositions *on, through, via, round, past, towards,* and *across* are used. The **Path** frame element is also expressed as the **Direct Object** for one verb, *come*.

In a second example (not shown), 52 lines were generated for the **Cure:Treatment** combination from a single instance of **through**, via the verb *rehabilitate.v* (sense 12, labeled **Intermediary** by the lexicographer, but essentially a **means** semantic role). The **Cure:Treatment** pair occurs in a much greater range of lexical items, including not only verbs (*alleviate, cure, ease, heal, rehabilitate, resuscitate,* and *treat*), but also nouns (*cure, healer, palliation, remedy, therapist, therapy,* and *treatment*) and adjectives (*curative, palliative, rehabilitative,* and *therapeutic*). Examining just those with a **Phrase Type** of PP, we see that *by, with, without*, and *for* are other prepositions in addition to **through** expressing the **Treatment** frame element.

Using the frames and frame elements from all sense-tagged instances as seeds, 9309 lines and 5440 lines similar to those in Table 5 are generated for **by** and **through**, respectively. (These files are also available in a tab-separated text file.) These results can be examined by sense number and can lead to an identification of all other prepositions expressing the frame elements as shown in Table 3. These prepositions are shown in Table 6 alongside those the lexicographer listed on the basis of intuition and Quirk assessments of semantic similarity.

The number of other prepositions expressing frame elements encompassed by a single sense was quite surprising. The first explanation for this large number was simply that the lexicographer had overlooked some possibilities. And indeed, upon reviewing the lists, the lexicographer could imagine substituting some of the suggestions in example sentences. However, the large number requires a more systematic explanation.

Table 6. Other Similar Prepositions for Senses of 'through'

| Sense | Lexicographer Prepositions | Prepositions Identifiable from FrameNet |
|---|---|---|
| 2 (1a) | into | into; on; over; about; at; across; in; under; against; between; through; around; with; behind; off; onto; towards; by; down; outside; along; near; below; beneath; above; of; within; underneath; beside; beyond; throughout; close; up; for; from |
| 3 (1b) | among, within | inside; through; under; within; at; beneath; amongst; between; on; behind; among; above; around; over; all; close; across; along; down; towards; up; past; via; from; of; alongside; by; with; to |

To assess the substitutability of other prepositions for a given semantic role, the lexicographer first examined their definitions in ODE for similarity. Many had similar definitions, but many did not. The lexicographer then examined the definitions in the *Oxford English Dictionary* (OED), which has a much larger number of senses than ODE. Rather than finding similar senses, the lexicographer concluded that, in fact, ODE simply provided a better organization of the many senses, ignoring obsolete and dated senses.

An immediate explanation for the large number of prepositions is simply to posit that prepositions are inherently polysemous. But, this seems to be too profligate a position. Instead, it seems much more likely that some meaning component of the attachment point (usually a verb) combines with some meaning component of the preposition to instantiate a frame element.

Instead of attempting to reach a final conclusion on substitutability, this issue will await further data when the other prepositions undergo their sense tagging. The analysis at that time will examine the semantic role assignments for prepositions deemed substitutable and determine their congruence. In particular, it will be possible to examine the array of frame elements of putative substitutable senses.

In addition to the other preposition analysis, the FrameNet data support an in-depth examination of other methods of realizing frame elements. For example, the alternation patterns for expressing the **Treatment** frame element appear to vary by part of speech of the lexical item. For verbs, we have "Comp PPing" (a complement prepositional phrase containing a gerund), "Ext NP" (an external argument, i.e., the subject of the verb), "DNI" (a definite null instantiation, indicating that the element is an anaphor), and a "Comp AVP" (a complement adverbial phrase, e.g. *treated pharmacologically*). Similar variations are indicated for nouns and adjectives. These semantic role alternations await further study.

## 6 Network of Preposition Senses

Litkowski (2002) claimed that prepositions in NODE could be arranged in a hierarchy based on a digraph analysis of the definitions. Prepositions do not seem a likely candidate for inheritance as in the case of nouns and verbs. The lexicographer examined this possibility in other preposition definitions ending in **by** (18) and **through** (6). Most cases using **by** (many of which included the phrase "supported by") did not seem to have a strong sense of inheritance, judged by the lexicographer as having an **Agent** sense based simply on the presence of the past participle.

The lexicographer also examined the 2-level hierarchy with ODE senses (core senses and their subsenses). ODE states that subsenses are usually generalizations or specializations of the core sense. In this effort, the lexicographer found that the relation of the subsenses to the core senses was based on some small bit of expanded or narrowed meaning. Whether these bits of meaning are involved in any putative inheritance will be studied further as TPP continues.

## 7 Conclusions and Further Work

The disambiguation of prepositions using a well-developed sense inventory and FrameNet instances has provided a wealth of data about the behavior of prepositions and semantic roles. Even though only two prepositions have been analyzed (at the time of submission), the results achieved extend across many semantic roles, numerous other prepositions, and semantic role alternation patterns. With only a modest amount of effort for disambiguating hundreds of instances, several large databases have been generated for further characterization of preposition and semantic role behavior. All data generated in The Preposition Project will be publicly available for researchers and application developers.[7]

The focus of The Preposition Project so far has been on establishing a framework for generating data and making it available. An important future part of the project will be in attempting to link this work with other research on prepositions (e.g., O'Hara & Wiebe, 2003 and Saint-Dizier, 2005).

The Preposition Project demonstrates considerable benefit available from exploiting the FrameNet databases. While the initial focus has been on preposition behavior, the semantic role alternations suggest the value of the FrameNet data for paraphrase opportunities.

---

[7]TPP files include the text file used to create the Excel instance spreadsheet, the Excel instance spreadsheet with sense tags, and the Excel sense analysis spreadsheet, and the summary lexicographic treatment of the preposition in a Word document.